# Next-generation reservoir computing validated by classification task

Ken-ichi Kitayama[*]

National Institute of Information and Communications Technology,
Tokyo, 184-8795, Japan
Hamamatsu Photonics K.K., Hamamatsu, 434-8601, Japan

* Corresponding author. email: kitayama@ieee.org

**SUMMARY** An emerging computing paradigm, so-called next-generation reservoir computing (NG-RC) is investigated. True to its namesake, NG-RC requires no actual reservoirs for input data mixing but rather computing the polynomial terms directly from the time series inputs. However, benchmark tests so far reported have been one-sided, limited to prediction tasks of temporal waveforms such as Lorenz 63 attractor and Mackey-Glass chaotic signal. We will demonstrate for the first time that NG-RC can perform classification task as good as conventional RC. This validates the versatile computational capability of NG-RC in tasks of both prediction and classification.
*key words: Reservoir computing, Prediction and classification*

## 1. Introduction

Reservoir computing (RC) is a unified computational framework [1-3], independently proposed recurrent neural network (RNN) models of echo state networks (ESNs) [4],[5] and liquid state machines (LSMs) [6],[7]. The primary difference from deep learning or multi-layer neural networks is that only the connections between the reservoir and the output layer are trainable, and the training requires much less data and computation than in the deep learning. Owing to the memory capability of the recurrent nature, it has been shown that RC can be preferably applied to prediction and recognition of temporal and sequential data such as spoken words [6], time series signals [5],[7], and wireless and optical channel equalizations [8],[9].

Recently, a new RC paradigm, named after "next-generation reservoir computing (NG-RC)" has been proposed [10], followed by several articles relevant to this topic [11]-[14]. True to its namesake, NG-RC requires no actual reservoir for input data mixing but rather computes polynomial terms directly from the inputs. In Figure1 both the architectures conventional RC and that of NG-RC are illustrated. The input temporal waveforms on x-, y-, and z-axes and the output classes are assumed. In NG-RC reservoir feature vectors are directly generated from the time series input, which include the polynomial terms of linear and nonlinear monomials. The choice of polynomial orders is flexible and task-dependent.

As for the optical hardware implementations, photonic frequency-multiplexed technique has been experimentally demonstrated [15]. Another optical implementation uses laser speckle patterns [16]. Benchmark tests so far reported, however, are limited to only prediction tasks of temporal waveforms such as Lorenz 63 attractor [10]-[14] and Mackey-Glass chaotic signal [12]. There has been a study of classification with NG-RC, which is relevant to a task of detecting specific three classes of anomaly in electrocardiography [17]. To validate the versatility of classification capability, an extensive study of classification tasks in NG-RC model will be needed , in which the outputs are several classes using a large time series dataset.

In this article, we will demonstrate for the first time that NG-RC can perform classification task as good as conventional RC having 1200-node reservoir. Our findings are that the performance critically depends on the combinations of linear and nonlinear monomials of time series input data.

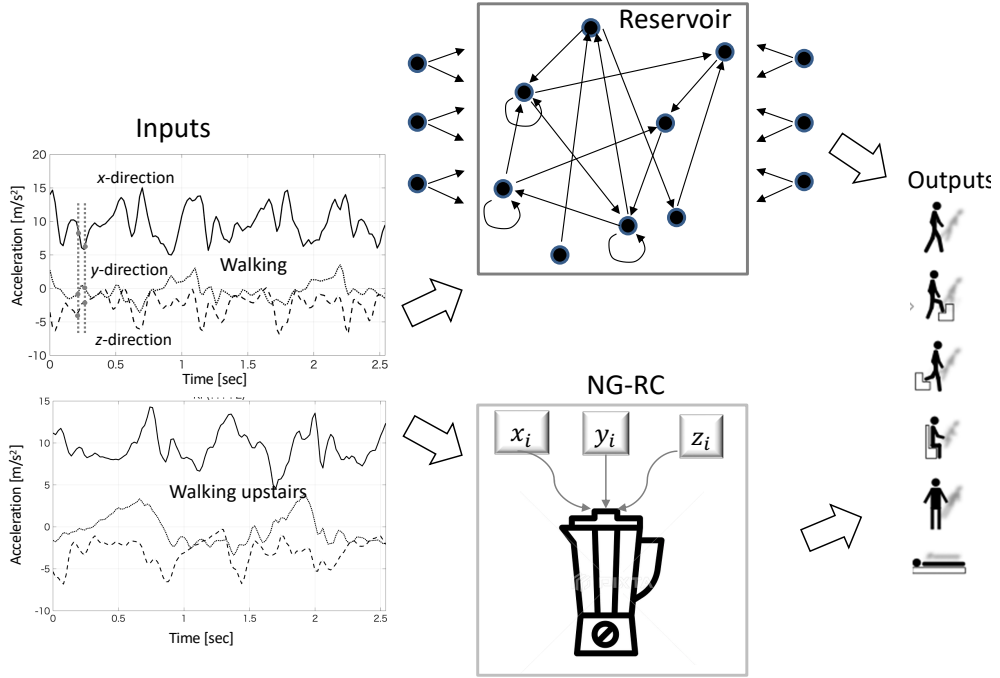

**Fig.1** Conventional RC vs. NG-RC. Conventional RC on the top processes time series input data on x-, y, and z-axes using an recurrent neural network. By contrast NG-RC at the bottom without physical nodes instead of nonlinear feature vectors of linear and nonlinear feature vectors, generated by mixing time-delay input time series data.

## 2. Next-Generation Reservoir Computing Model

To begin with, let us assume $\mathbb{r}_i$ at discrete time step $i$ $(i = 1,2,\cdots)$, the linear and nonlinear feature vectors at the time step $i$ $(i = 1,2,\cdots)$ are expressed by

$$\mathbb{O}_{lin} = \mathbb{r}_i \oplus \mathbb{r}_{i-1} \cdots \oplus \mathbb{r}_1 \quad (1)$$

$$\mathbb{O}_{nl}^{(p)} = \mathbb{O}_{lin} \otimes \mathbb{O}_{lin} \otimes \cdots \otimes \mathbb{O}_{lin} \ (p = 1,2\cdots) \quad (2)$$

where $\oplus$ is vector concatenation operation, $\otimes$ denotes product, and $p$ represents the order of the polynomial. By using the above feature vectors $\mathbb{O}_{lin}$ and $\mathbb{O}_{nl}^{(p)}$, the total feature vector $\mathbb{O}_{total}$ is given by

$$\mathbb{O}_{total} = \mathbb{O}_{lin} \oplus \mathbb{O}_{nl}^{(p)} \quad (3)$$

Here we assume that the weight ratio of $\mathbb{O}_{nl}^{(p)}$ to $\mathbb{O}_{lin}$ is unity. The output corresponds to discrete class label of human activity in Figure 1, which includes six human motions, "walking", "walking upstairs", "walking downstairs", "sitting", "standing", and "laying". The desired output $\mathbb{Y}$ is obtained with $\mathbb{O}_{total}^{test}$ of test dataset by

$$\mathbb{Y} = W_{out} \mathbb{O}_{total}^{test} \quad (4)$$

The output weight $W_{out}$ is uniquely determined by employing the least squares method by using the regularization parameter $\lambda$ as

$$W_{out} = \mathbb{Y}_{target} \mathbb{O}_{total}^{tr\,T}(\mathbb{O}_{total}^{tr} \mathbb{O}_{total}^{tr\,T} + \lambda I_{N\times N})^{-1} \quad (5)$$

where $\mathbb{Y}_{target}$ reperesents the target output to $\mathbb{O}_{total}^{tr}$ of the training dataset. Please note that the total feature vector $\mathbb{O}_{total}$ in Equations (4) and (5) are distinct, representing those of the test dataset and the training dataset, respectively. The target output $\mathbb{Y}_{target}$ in equation (5) is expressed by a one-hot vector, in which each column represents a class, e.g. the number of classes $S = 6$;

$$\mathbb{Y}_{target} = \begin{bmatrix} 1 & 0 & 0 & 0 & 0 & 0 \\ 0 & 1 & 0 & 0 & 0 & 0 \\ 0 & 0 & 1 & 0 & 0 & 0 \\ 0 & 0 & 0 & 1 & 0 & 0 \\ 0 & 0 & 0 & 0 & 1 & 0 \\ 0 & 0 & 0 & 0 & 0 & 1 \end{bmatrix} \quad (6)$$

**Table 1** *Polynominal terms generated by 128-time series input data on x-, y-, and z-axes, $\mathbb{r}_i\ (i = 1, \cdots 128)$*

| | Abbreviations | Monomials | Total numbers |
|---|---|---|---|
| | | | 2423 |
| Linear $\mathbb{O}_{lin,i}$ | $lin$ | $x_i, x_{i-1}, x_{i-2}, \cdots$ <br> $y_i, y_{i-1}, y_{i-2}, \cdots$ <br> $z_i, z_{i-1}, z_{i-2}, \cdots$ | 384 <br> $(= 128 \times 3)$ |
| Nonlinear $\mathbb{O}_{nl,i}^{(p)}$ | $nls$ <br> $(p = 2)$ | $x_i x_{i-1}, x_{i-1}x_{i-2}, \cdots$ <br> $y_i y_{i-1}, y_{i-1}y_{i-2}, \cdots$ <br> $z_i z_{i-1}, z_{i-1}z_{i-2}, \cdots$ | 381 <br> $(= 127 \times 3)$ |
| | $nlq$ <br> $(p = 2)$ | $x_i x_i, x_{i-1}x_{i-1}, \cdots$ <br> $y_i y_i, y_{i-1}y_{i-1}, \cdots$ <br> $z_i z_i, z_{i-1}z_{i-1}, \cdots$ | 384 |
| | $nlcs$ <br> $(p = 2)$ | $x_i y_{i-1}, x_{i-1}y_i, \cdots$ <br> $y_i z_{i-1}, y_{i-1}z_i, \cdots$ <br> $z_i x_{i-1}, z_{i-1}x_i, \cdots$ | 762 |
| | $nlt$ <br> $(p = 3)$ | $x_i y_i z_i, x_{i-1}y_{i-1}z_{i-1}, \cdots$ | 128 |

Let us consider the time series input data on x-, y-, and z-axes, $\mathbb{r}_i = (x_i, y_i, z_i)$. The dataset of human activity in Figure 1 was used in the present benchmark test of classification task, which includes six human motions, "walking", "walking upstairs", "walking downstairs", "sitting", "standing", and "laying". Polynomials of the feature vector $\mathbb{O}_{total}$ are listed up in Table 1, including monomials of the linear $\mathbb{O}_{lin}$ and nonlinear $\mathbb{O}_{nl}^{(p)}$ ($p = 2,3$). *lin* represents the linear monomial, and *nls, nlq, nlcs,* and *nlt* are the nonlinear monimials of time-delayed product, representing the square with time-delay, square without time-delay, cross-product without time-delay, cross-product with time-delay, and cross-cube without time-delay. The total number of monomials is 2039. Although the nonlinear polynomials could include a wide variety of monomials of the higher-order terms as much as desired. The vector $\mathbb{O}_{total}$ in Equation (3) is obtained by stacking column-wise all the polynomials of the feature vectors $\mathbb{O}_{lin}$ and $\mathbb{O}_{nl}^{(p)}$.

## 3. Benchmark Test: Classification Task

We conducted the simulation of benchmark test of classification of six human motions. The temporal waveforms along x-, y-, and z-axes were captured for 2.56 sec by an accelerator of smartphone and sampled into 128-sample on each axis at the rate of 20 ms [18]. An example of the waveforms of "walking" in the training dataset is shown in Figure 1. All the polymonials of the feature vectors of the linear $\mathbb{O}_{lin}$ and nonlinear $\mathbb{O}_{nl}^{(p)}$, ($p = 2,3$) listed in Table 1 are calculated by using the time series of $384(= 128 \times 3)$ sampled data in the numerical simulations. The datasets of the training and the testing include 7352 and 2947, respectively [18]. The nonlinear polynomials in Table 1 are calculated using the 7352 training dataset of and 2947 of testing test dataset. We used the raw dataset [18] where the magnitude of these three-dimensional signal was calculated using the Euclidean norm (L2). The second-order nonlinear polynomials, *nls, nlq,* and *nlcs* and the third-order *nlt* are calculated by using the temporal waveforms of six human motions.

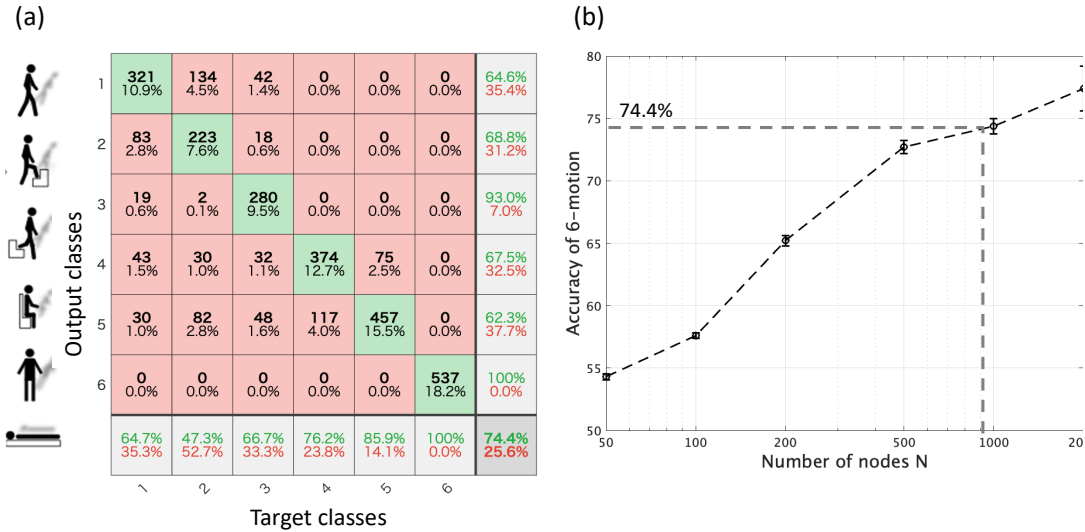

**Fig.2** (a) Confusion matrix of the accuracy of six human motions with uniform weight. (b) The result of conventional RC with (k, p) = (4,0.5) where k and p denotes the degree of node and the probability of rewiring, respectively, and the spectral radius is 8.41. The accuracy is plotted as a function of the number of nodes in the reservoir.

It is assumed that all the weights on the five monomials are uniform. The best performance is the accuracy of 74.4%. The confusion matrix of the best result is shown in Figure 2(a). Please note that the examinees of training data are different from those of the test data, resulting generally in lower accuracy than the case with the same examinees both in the training and test. The accuracy 74.4% of NG-RC is equivalent to the result of conventional RC with nearly 1000-node reservoir as shown in Figure 2(b). As far as the classification task is concerned, this result confirms that NG-RC can perform as good as conventional RC having a relatively large reservoir. Besides, the computation time is much shorter than conventional RC. The reservoir of conventional RC is based on the small-world network [19] having the node degree $k = 4$ and the probability of rewiring $p = 0.5$ [20]. The small-world network is categorized between a regular network ($p = 0$) and a completely disordered one ($p = 1$) where a small amount of the links between the nodes are rewired to introduce disorder. The node degree $k = 4$ indicates connections with $2k$ neighboring nodes. It is noteworthy that the value of spectral radius $\rho(W_{res})$ in this simulation with the conventional small-world based reservoir is typically around 8 where $W_{res}$ denotes the weight matrix of reservoir. According to [21], the echo state property holds for the spectral radius $\rho(W_{res}) < 1$, however, the echo state property often holds even when $\rho(W_{res}) \geq 1$. In fact, the present spectral radius can be significantly greater than 1.

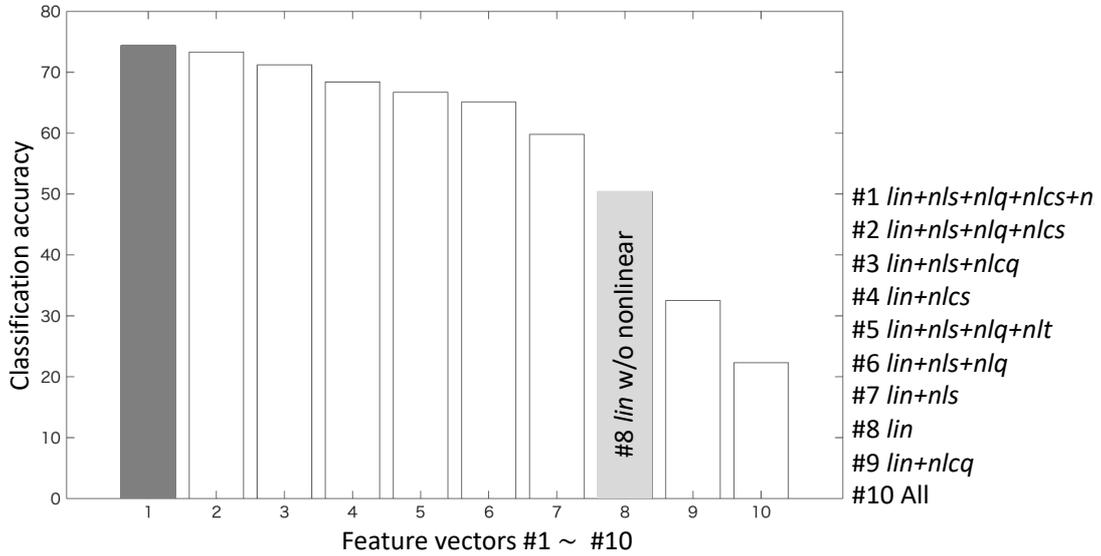

**Fig.3** The classification accuracies vs. feature vectors #1−#10. The best performance is shown on the far-left in dark gray, and the result with #8 lin without nonlinear polynomial in light gray. #1: *lin+nls+nlq+nlcs+nlt*, #2: *lin+nls+nlq+nlcs*, #3: *lin+nls+nlcq*, #4: *lin+nlcs*, #5: *lin+nls+nlq+nlt*, #6: *lin+nls+nlq*, #7: *lin+nls*, #8: lin, #9: *lin+nlcq*, and #10: all, *lin+nls+nlq+nlcq+nlcs+nlt*.

We investigate impact of various feature vectors on the results of benchmark test. In Figure 3 the classification accuracies for #1−#10 feature vectors $\mathbb{O}_{total}$ are compared. The performance critically depends on the combinations of linear and nonlinear monomials of time series input data. The best performance of the accuracy of 74.4% is obtained with the feature vector of #9 shown in dark gray. The accuracy with #10 feature vector including all the monomials is unexpectedly the worst 22.3%, which is even worse than 50.4% of #8 with only linear feature vector $\mathbb{O}_{lin}$ in light gray. Noteworthy is that the improvement in the accuracy from #2 to #1 is observed by adding even the third-order polynomial *nlt*.

There remains opportunity to improve further the performance by tailoring weights on the above five nonlinear polynomials. As shown in Figure 4 the accuracy went up from 74.4% with uniform weight in Figure 2 to 75.4% with the following nonuniform weights ; $(w_{lin}, w_{nls}, w_{nlq}, w_{nlcs}, w_{nlt}) = (1.0, 1.8, 2.0, 1.4, 0.4)$. This amounts to that of conventional RC having roughly 1200-node in the conventional reservoir. It should be pointed out that the methodology for the NG-RC of the best performance has not yet been established how to determine the weights on polynomial terms, and this problem remains an open question.

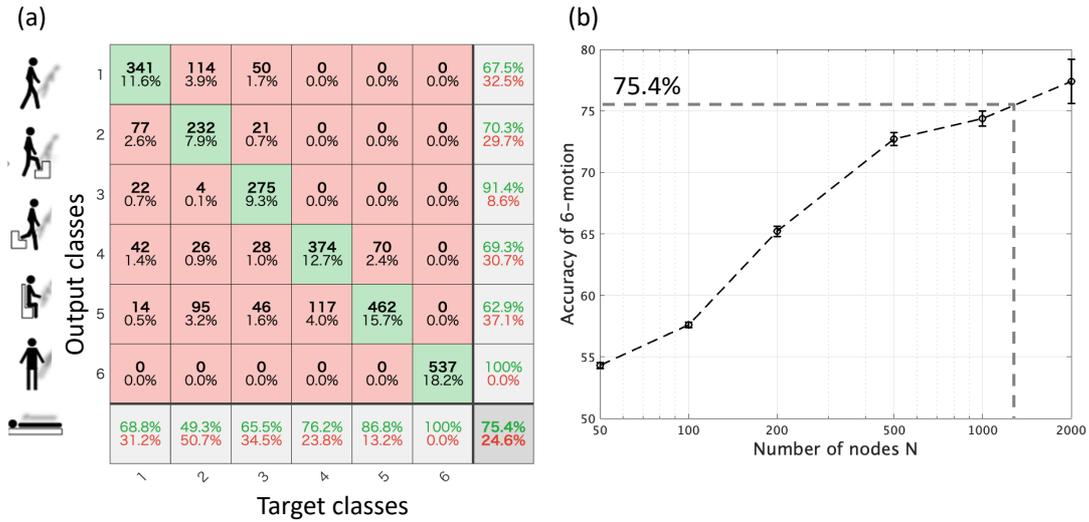

**Fig.4** (a) Confusion matrix of the accuracy of six human motions with nouniform weight, $(w_{lin}, w_{nls}, w_{nlq}, w_{nlcs}, w_{nlt}) = (1.0, 1.8, 2.0, 1.4, 0.4)$. (b) The result of conventional RC with $(k, p) = (4, 0.5)$ where k and p denotes the degree of node and the probability of rewiring, respectively, and the spectral radius is 8.41. The accuracy is plotted as a function of the number of nodes in the reservoir.

## 4. Conclusion

We have demonstrated for the first time that NG-RC can perform classification task as good as conventional RC having 1200-node reservoir. This validates the versatile computational capability of NG-RC in tasks of both prediction and classification. An important finding was that the performance critically depends on weighing the linear and nonlinear monomials.

## Acknowledgments

I would like to thank for valuable discussions H. Harai, Y. Awaji, and H. Furukawa of the National Institute of Information and Communications Technology, H. Toyoda and the members of AI Working Group of Hamamatsu Photonics Central Laboratory.

## References

[1] D. Verstraeten, B. Schrauwen, D. Stroobandt, and J. Van Campenhout, "Isolated word recognition with the liquid state machine: A case study," *Inf. Process. Lett.* 95, pp.521–528, 2005.
[2] M. Lukoševicius and H. Jaeger, "Reservoir computing approaches to recurrent neural network training," Comput. *Sci. Rev.* 3, pp.127–149, 2009.
[3] M. Yan, C. Huang, P. Bienstman, P. Tino, W. Lin, and J. Sun, "Emerging opportunities and challenges for the future of reservoir computing," *Nature Commun* 15:2056, 2024.
[4] H. Jaeger, "The 'echo state' approach to analyzing and training recurrent neural networks," *Technical Report GMD Report* 148, German National Research Center for Information Technology, 2001.
[5] H. Jaeger and H. Hass, "Harnessing nonlinearity: Predicting chaotic systems and saving energy in wireless communication," *Science* 304, pp.78–80, 2004.
[6] W. Maass, T. Natschläger, and H. Markram, "Real-time computing without stable states: A new framework for neural computation based on perturbations," *Neural Comput.* 14, pp.2531–2560, 2002.
[7] W. Maass, "Liquid state machines: Motivation, theory, and applications," In *Computability in Context: Computation and Logic in the Real World*, pp.275–296 (World Scientific, 2011).
[8] P. Antonik, F. Duport, M. Hermans, A. Smerieri, M. Haelterman, and S. Massar, "Online training of an opto-electronic reservoir computer applied to real-time channel equalization," *IEEE Trans. Neural Networks and Learning Sys.*, vol.28, No.11, pp.2686–2698, 2017.
[9] A. Argyris, J. Bueno, and I. Fischer, "Photonic machine learning implementation for signal recovery in optical communications," *Sci. Rep.*, 8, 8487, 2018.
[10] D. J. Gauthier, E. Bollt, A. Griffith, W. A. S. Barbosa, "Next generation reservoir computing," *Nature Commun*, 12: 5564, 2021.
[11] R. Pyle, N. Jovanovic, D. Subramanian, K. V. Palem, and A. B. Patel,, "Domain-driven models yield better predictions at lower cost than reservoir computers in Lorenz systems," *Philosophical Transactions of the Royal Society A*, 379, 20200246, 2021.
[12] E. Bollt, "On explaining the surprising success of reservoir computing forecaster of chaos? The universal machine learning dynamical system with contrast to VAR and DMD," *Chaos: An Interdisciplinary Journal of Nonlinear Science*, 31, 2021.
[13] W. A. Barbosa and D. J. Gauthier, "Learning spatiotemporal chaos using next-generation reservoir computing," *Chaos: An Interdisciplinary Journal of Nonlinear Science*, 32, 2022.
[14] L. Grigoryeva, H. L. J. Ting, and J.-P. Ortega, "Infinite dimensional next-generation reservoir computing," *arXiv*:2412.09800v3 [cs.LG] 21 Feb 2025
[15] N. Cox, J. Murray, J. Hart, and B. Redding, "Photonic frequency multiplexed next-generation reservoir computer," *APL Photon*. 10, 036122, 2025.
[16] H. Wang, J. Hu, Y. Baek, K. Tsuchiyama, M. Joly, Q. Liu, and S. Gigan, "Optical next generation reservoir computing," *arXiv*:2404.07857v3, 2024.
[17] K. Arbateni, and A. Benzoui, "Enhancing heartbeat classification throughcascading next generation and conventional reservoir computing." *Applied Sciences* 14, 3030, 2024.
[18] D. Anguita, A. Ghio, L. Oneto, X. Parra, and J. L. Reyes-Ortiz, "Activity recognition on smartphones using a multiclass hardware-friendly support vector machine," *International Workshop of Ambient Assisted Living* (*IWAAL*2012) (Vitoria-Gasteiz, Spain, December 2012).
[19] S. H. Strogatz and I. Stewart, "Collective dynamics of 'small-world' networks," *Nature*, 393, 440–442, 1998.
[20] K. Kitayama, "Guiding principle of reservoir computing based on 'small-world' network," *Sci. Rep.*, 12 16697, 2022.
[21] M, Lukoševičius, "A practical guide to applying echo state networks," *Neural Netw. Tricks Trade* 20, 659–686, 2012.